# Loss convergence in a causal Bayesian neural network of retail firm performance

F. Trevor Rogers[a]

## Abstract

We extend the empirical results from the structural equation model (SEM) published in the paper Assortment Planning for Retail Buying, Retail Store Operations, and Firm Performance [1] by implementing the directed acyclic graph as a causal Bayesian neural network. Neural network convergence is shown to improve with the removal of the node with the weakest SEM path when variational inference is provided by perturbing weights with Flipout layers, while results from perturbing weights at the output with the Vadam optimizer are inconclusive.

## Introduction

Structural equation models (SEM) have been implemented as neural networks in a variety of domains, such as sustainability [2]. Furthermore, graphs are used in retail by Amazon.com for establishing relationships between items [3]. This research follows by implementing a structural equation model from merchandising research as a directed acyclic graph neural network that is causal in nature. Weight perturbation is used in the layers of the graph nodes to provide probabilistic Bayesian uncertainty estimates of retail firm performance.

Research efforts are underway to incorporate SEM directly into neural networks, in methods such as DAG-GNN [4]. Statisticians have also used neural network optimizers to estimate the covariance matrices of SEM models in order to provide regularization, in software such as TensorSEM [5]. However, there is still something to be gained by using SEM separately to validate the causal relationships of a directed acyclic graph before training it as a neural network.

Specifically, this research demonstrates that removing the graph node with the weakest SEM path coefficient improves the convergence of a causal graph neural network. The path model under study improved convergence by 62.5% when the weakest node from the structural equation model was removed from the graph neural network (see Figure 1).

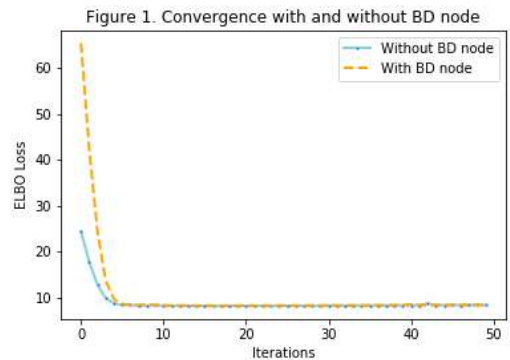

Figure 1. Convergence improves with the removal of the node with the weakest SEM path when Flipout layers are used.

---

[a] rogers72@hawaii.edu.

In this manner, the path analysis provided by SEM validated the assumptions that directed acyclic graphs make about conditional independence between graph nodes, improving training. In turn, the Bayesian graph neural network can reciprocate by exploring the latent space of the model, such that the process of hyperparameter tuning helps to resolve issues of factor (feature) indeterminacy that emerge in SEM when the path analysis performs equally well with and without some of the features that contribute to a node.

**From SEM to neural network**

The graph structure of the data in this study is expressed by a pair of nested equation. The first describes the residual connections of the buyer demographics (BD) and store management (SM) nodes as they inform the assortment success (AS) node (with error). Equation (2) describes how (1), SM, BD and error all affect firm performance (FP):

$$AS = {}_{pathBD\text{-}AS}BD + {}_{pathSM\text{-}AS}SM + e1_\epsilon \quad (1)$$

$$FP = AS + {}_{pathSM\text{-}FP}SM + {}_{pathBD\text{-}FP}BD + e2_\epsilon \quad (2)$$

The relationship between these nested equations can be seen in Figure 2. In terms of graph structure, the SEM model constitutes two V-structures of nodes, or collider cases, that also share nodes in what is termed an 'immorality.' In both V-structures, SM and BD both explain AS, while firm performance is also explained by SM and BD as well as AS.

These nested equations together form (3), the observed covariance matrix, which in turn is used with a model-implied covariance matrix (4) for SEM. Using the standard Jöreskog-Keesling-Wiley LISREL notation for SEM, these equations are expressed as:

$$y = X\eta + \epsilon \quad (3)$$

$$\eta = \beta_0\eta + \xi \quad (4)$$

where $X$ and $y$ are the observed parameters of (1) and (2), $\epsilon$ and $\xi$ are uncorrelated error terms, and $\eta$ is an endogenous random vector of interest for which SEM estimates $\beta_0$, the regression path weights in Figure 3, which shows weaker weights for the BD-AS and BD-FP paths.

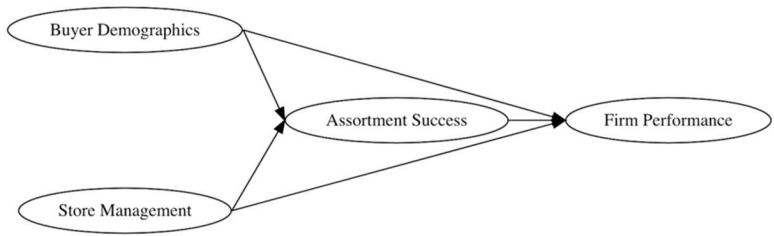

Figure 2. The firm performance mode showing the paths from buyer demographics (BD) and store management (SM) to assortment success (AS) and then to firm performance (FP).

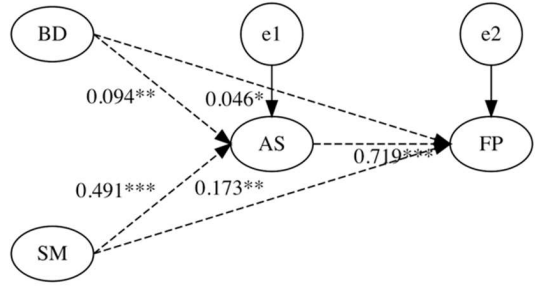

Figure 3. The structural equation model showing the paths from buyer demographics (BD) and store management (SM) to assortment success (AS) and then to firm performance (FP). The BD paths are the weakest path in the model. *** denotes statistical significance at the 0.001 level, ** = 0.01, and * = 0.1.

When $\eta$ is probabilistic in nature, SEM becomes a Bayesian network [6], such that $y$ is now a probability distribution $P(x)$ on $X$ such that $P(x)$ for all x sums to 1. The conditional probability for a single parent-child node of such a causal Bayesian network relationship is expressed with the summation:

$$\sum_{x_i \in X} P(x_i | pa(x_{(i)})) = 1 \qquad (5)$$

where $\beta_i$ becomes a vector of probabilities over all values of $x_i$ from parent nodes $pa(x_{(i)})$ instead of a single regression path coefficient in SEM. When the causal Bayesian network becomes a causal Bayesian graph neural network, then using the graph network notation conventions of Bayesian neural networks from [7], the network is expressed as the product of the summations of (5) according to the density of the neurons in the parent node, creating a joint probability for all $x_i$:

$$P(x_1, \ldots, x_n) = \prod_{i=1}^{n} P(x_i | pa(x_{\pi(i)})) \qquad (6)$$

For $x \in \mathbf{R}^n$, where R is the feature vector for node $n$ in graph G. (6) is estimated at each node in Figure 2, together predicting the uncertainty in the posterior distribution given the causal relationship between the nodes that has been validated by SEM.

As a preliminary to the SEM of the graph that validates the causal Bayesian neural network, factor analysis is conducted to reveal both eigenvalues as well as the percentage of variance explained of the variables, or the communality (see Table 1). Both of these measures denote the quality of the nodes in relation to each other, apart from the final path coefficients from estimating the structural equations in Figure 3. The AS and FP factors have two values each in table 1 since they were measured both in relation to the firms under study as well as their competitors.

| Factor group | Features per node | Eigenvalue | % of variance |
|---|---|---|---|
| AS (compared to sales plan) | 6 | 3.24 | 22.41 |
| AS (compared to competitors) | 6 | 1.0 | 23.27 |
| FP (compared to sales plan) | 5 | 0.73 | 24.21 |
| FP (compared to competitors) | 5 | 0.61 | 22.04 |
| Retail store management (SM) | 9 | 0.23 | 8.05 |
| Buyer demographics (BD) | 39 | 0.182 | 0.03 |

Table 1. Factor analysis results again showing the weakness of the BD node in the eigenvalues as well as the proportion of the variance in FP explained by that node.

Again, as also seen from the path weights of Figure 3, the eigenvalues of the BD node are much weaker than the other nodes. Because of this, the neural network is trained with and without BD to determine the effect of its removal.

**Neural network**

The path weights of the graph in Figure 3 are derived using R's Lavaan package for structural equation modeling. Here, the graph is implemented as a neural network using version 0.7.0 of Tensorflow-probability. Flipout layers are used that perturb weights with a random Rademacher and the Kullback-Leibler divergence to produce probabilistic uncertainty estimates [8]. For comparison, a variant of Kingma and Ba's Adam optimizer is also used [9]. The Vadam optimizer perturbs weights without the KL divergence at the output layer [10].

To create a directed acyclic graph neural network from the structural equations, a neural network is created for each of equations (1) and (2). The two V-structured networks are combined together in a connected double-chevron architecture, where the output of (2) is the input of (3) and the same batch-wise inputs from SM and BD are used simultaneously in each neural network during training. Both neural networks also have their own loss function and optimizer, the losses of which are summed for the final output layer. The loss functions for both V-structures are the negative log likelihood losses to which the Kullback-Leibler divergences from the Flipout layers in each V-structure are added to form the evidence lower-bound (ELBO) loss. Both KL losses are scaled by $1/n$, where n is the normalizing constant of the number of observations in training data.

In addition, some changes are made to the graph in Figure 1. Extra Flipout layers are added to accommodate learning wherever there is input from the BD features since nominal categorical encoding required that those 12 features be expanded to 39 features. In addition, the inputs of SM and BD are concatenated together in the first V-structure, and again with the output of that V-structure to predict firm performance. Concatenation is used following guidance on combining inputs in graph neural networks from [11], where performance improved using concatenation layers instead of merging layers. Finally, because the five firm performance predictors are all ordinal and similarly scaled, the linear combination of those predictors is used for the sake of computational simplicity.

The distribution of FP values roughly follows a Weibull distribution with some missingness one and two standard deviations above the mean, but a Gaussian distribution performed better in practice. Therefore, the output layers for each neural network are non-negative Gaussian distributions with a learnable tuning parameter that affects the mean and the standard deviation of the output distribution, which is initially mean centered at 25, the midpoint of the absolute ordinal scale for FP. The resulting neural network is tuned using the Hyperas wrapper for the Hyperopt hyperparameter tuning library [12].

In both V-structures of the causal graph neural network, the input size is used as the central hyperparameter value for layer density with two higher and lower values added to the search space, offering the potential for layers that are denser or narrower than the input by several neurons. The choices made by Hyperas during the tuning process reveal information about the factor indeterminacy of the SEM graph model that are not revealed by conventional statistical methods. If Hyperas determines that the optimal node density is wider than the feature input, that indicates greater aleotoric uncertainty and possible missing latent variables in the graph structure, resolving issues of factor indeterminacy in the SEM. In addition to node density, the Adam optimizers are also tuned by learning rate with ranges between 0.01 and 0.1, where values just greater than 0.01 are chosen. ReLU are used by default. Figure 4 has the architecture chosen by Hyperas.

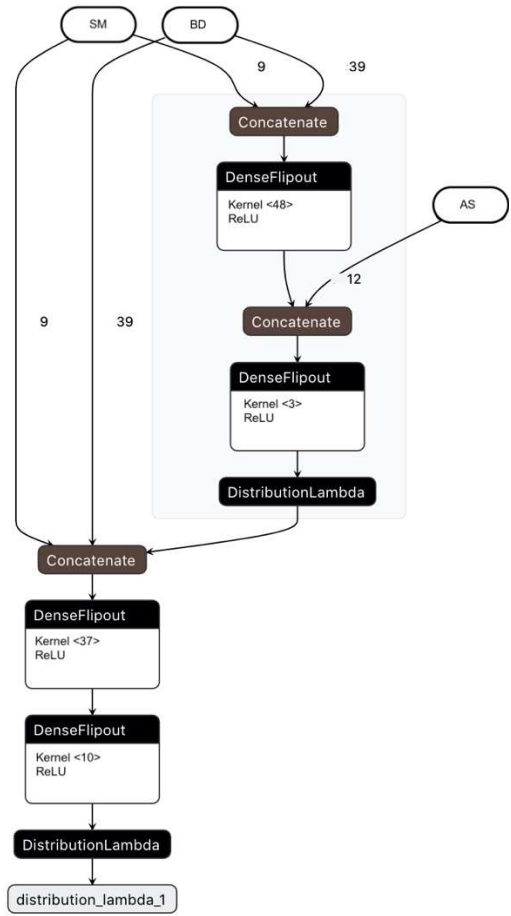

Figure 4. The neural network architecture chosen by Hyperas

**Node Permutation**

The choices in Figure 4 indicate that for both neural networks, the chosen input layer widths are narrower than the input sizes. This indicates less aleotoric uncertainty in the data and that additional features are not necessary to predict retail firm performance. This architecture was subsequently trained for 50 epochs with 10 repetitions of 10-fold repeated holdout validation and a batch size of 4.

To test the factor indeterminacy suggested by the structural equation model, the architecture was trained without the weakest node by path weights, the BD node. To test the variational inference provided by KL divergences from the weight perturbation in the Flipout layers, alternate neural networks are specified that replace the Flipout layers with normal dense layers and replace each of the Adam optimizers with Vadam optimizers that perturbs weights at the output layer instead of in the middle layers. Vadam is scaled using the training set size, and uses a prior distribution precision of 0.1, a slightly informative prior precision used in the ablation tests of [6]. It was trained using one Monte Carlo sample to match Flipout, though more samples are advised.

**Results**

Table 2 reports the lowest losses from the repeated holdout cross-validation with Flipout. The results show almost no difference in loss from the removal of any single node. However, Figure 5 shows that the model converged to a solution more quickly when the BD node was removed, with initial epoch loss dropping 62.5%.

| Results | | | | |
|---|---|---|---|---|
|  | Full model | No BD | No SM | No AS |
| Flipout | 8.19 | 8.15 | 8.03 | 8.04 |
| Vadam | 10.05 | 10.41 | 10.20 | 10.25 |
| Both | 7.73 | 7.73 | 7.72 | 7.73 |

Table 2. Reported validation loss as represented by the predicted tuning parameter for the mean and standard deviation of the output probability distribution.

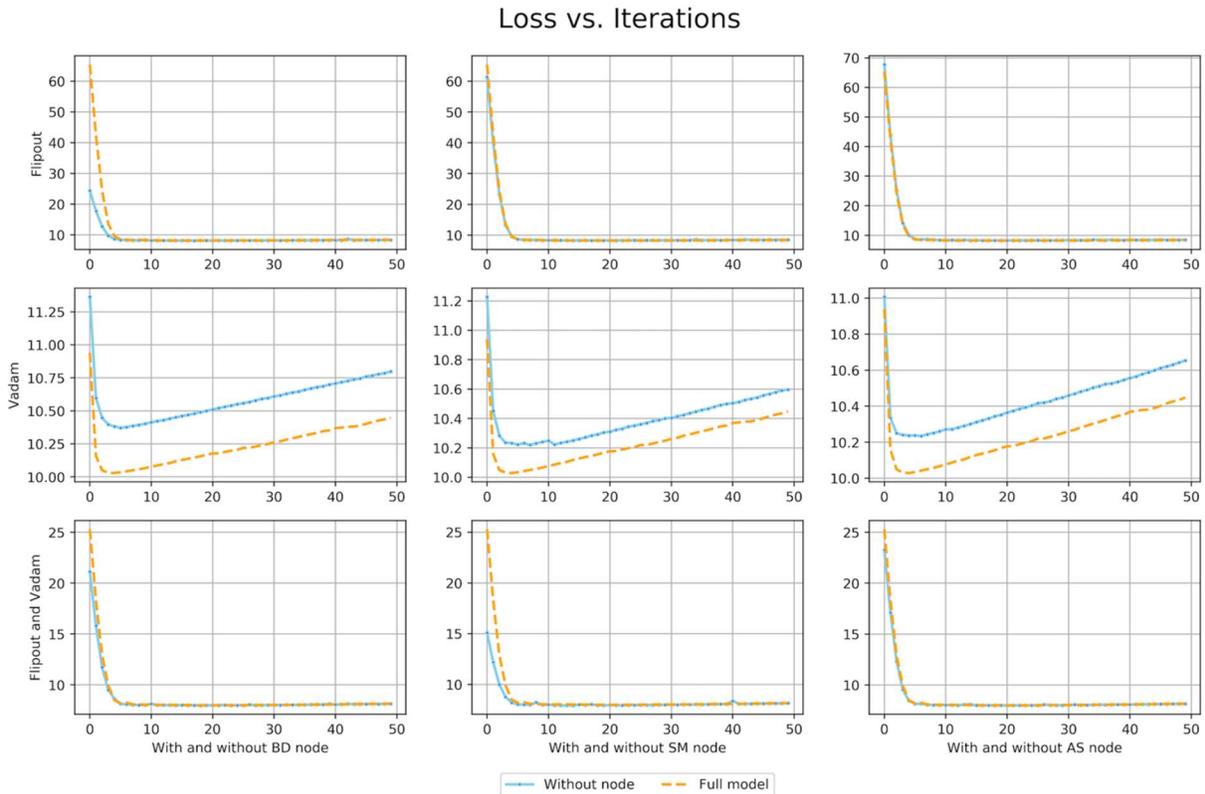

Figure 5. Loss over 50 iterations for the results from table 2. ELBO loss is reported for Flipout and Flipout/Vadam. The upper-left plot is Figure 1.

This indicates that the benefits of removing the node with the weakest SEM path weights are in the improvement of loss convergence.

For comparison, Figure 5 also shows the losses of the neural network when the SM or AS nodes are removed from the neural network against the loss of the full network, and when these losses are from the use of Flipout, the use of Vadam, and the use of both Flipout and Vadam together. The proportional differences in convergence from the removal of the AS and SM nodes can immediately be seen for the main results using Flipout. The losses from the use of Flipout are also initially higher than that of the models estimated using Vadam. This can primarily be attributed to the Kullback-Leibler divergences from each layer that are added together with the total loss from the Flipout layers, forming the Evidence Lower-Bound loss (ELBO).

When the Vadam optimizer is used alone, the neural network converges quickly but loss increases immediately afterwards, and convergence suffers from the removal of the node instead of improving. This could largely be attributed to lower amount of variation added by Vadam, which only perturbs weights at the output layer and does so without the KL divergence and only with Monte Carlo sampling. Only one sample was used to match Flipout for comparison, and many more samples are clearly needed to provide weight perturbance comparable to that provided by Flipout.

Combined with Flipout layers, Vadam instead seems to work as a regularizer for the weight perturbance from the Flipout layers, and in comparison with the results using Flipout layers alone, this shows that removing the weakest path node found by SEM improves convergence in proportion to the amount of perturbance added to the network when the KL divergence is used. In this example, removing the node with the weakest probabilistic relationship to firm performance reduced the KL divergence and subsequently improved neural network convergence. For much deeper causal Bayesian graph networks that are modelling more complicated causal relationships, this could result in much larger convergence gains in proportion to the size of the network and to the amount of weight perturbance in that network, and perhaps also to the number of nodes removed from finding weak path connections.

**Future Research**

Though there are research efforts underway to use structural equation model-style combinatorics in neural networks, and concurrent efforts to use neural network optimizers to estimate the covariance matrices of SEM, there may still be value in using SEM separately before training a causal Bayesian neural network training to determine the strength of the assumed joint probabilities. Far from suggesting that SEM not be incorporated in to neural network training, this research suggests that SEM might be used during causal Bayesian graph neural network training at each iteration or epoch as a form of node-wise dropout that removes nodes during graph training based on the path coefficient estimated using weights. Future developments in neural network research using SEM will certainly hold developments such as this.